# Patch-Based Image Hallucination for Super Resolution with Detail Reconstruction from Similar Sample Images

Chieh-Chi Kao, Yuxiang Wang, Jonathan Waltman, Pradeep Sen

*Abstract*—Image hallucination and super-resolution have been studied for decades, and many approaches have been proposed to upsample low-resolution images using information from the images themselves, multiple example images, or large image databases. However, most of this work has focused exclusively on small magnification levels because the algorithms simply sharpen the blurry edges in the upsampled images – no actual new detail is typically reconstructed in the final result. In this paper, we present a patch-based algorithm for image hallucination which, for the first time, properly synthesizes novel high frequency detail. To do this, we pose the synthesis problem as a patch-based optimization which inserts coherent, high-frequency detail from contextually-similar images of the same physical scene/subject provided from either a personal image collection or a large online database. The resulting image is visually plausible and contains coherent high frequency information. We demonstrate the robustness of our algorithm by testing it on a large number of images and show that its performance is considerably superior to all state-of-the-art approaches, a result that is verified to be statistically significant through a randomized user study.

*Index Terms*—Image hallucination, image super-resolution, large image databases.

## I. INTRODUCTION

The problem of upscaling a small, bitmapped image to produce a larger, realistic image has many practical multimedia applications. Because of its importance, it has been well-studied for the past few decades under names such as image super-resolution, image upsampling/upscaling, or image hallucination.[1] Although upsampling the input image is straightforward with traditional interpolation schemes (e.g., [3]), the main challenge is to re-introduce high-frequency detail in a plausible manner.

Two general kinds of methods have been introduced to do this. First, *image super-resolution* methods aim to recover the true detail that would have been present in the theoretical high-resolution image. However, this is an ill-posed problem since that information has already been lost in the downsampling process. For this reason, blurry artifacts are often produced *in lieu* of the missing detail. On the other hand, *image hallucination* algorithms have the freedom to synthesize new detail that may diverge from the theoretical high-resolution image as long as they remain consistent with the thumbnail input provided. In theory, this would allow them to synthesize new high-resolution detail and produce high-quality results.

While image hallucination may not be suitable for scientific, medical, or military applications, it has a wide range of artistic uses from photography to advertising to personal use.

Many different kinds of algorithms have been proposed to upscale a small image to a larger one with high-frequency detail, from those that use statistical image priors (e.g., [4]) to those that draw detail from examples in an image database (e.g., [5, 6, 7]). However, despite this large amount of previous work, no image super-resolution/hallucination algorithm exists that can add significant new high-frequency detail to the input images in a robust and plausible manner. Even state-of-the-art methods mostly sharpen edges and other detail *that already exists* in the low-resolution image (see Fig. 1a–c). For example, given an input thumbnail of a scene with a tree that is only a few pixels in size, we are not aware of an existing algorithm that can upscale this and synthesize a plausible tree with distinct leaves that resembles the input image.

In this paper, we propose an image hallucination algorithm that can insert new, synthesized detail into upsampled images to produce plausible, high resolution results by using sample images. This allows us to explore a regime of more extreme image magnification than was studied in previous work, where image resolution was usually increased by a factor of $8\times$ or less (typically 2 to $4\times$). For example, as shown in Fig. 1d, our method allows us to turn a tiny, $64 \times 48$ thumbnail image into a plausible megapixel image (a magnification of $32\times$) with significant added detail.

This kind of method could be applied to many multimedia applications, such as restoring old, low-resolution photographs or transforming standard images into browsable, interactive experiences. Also, the method could be used to hallucinate images of people from low-resolution thumbnails when high-resolution examples are available, say from a personal photo collection (see Fig. 6). Furthermore, the method can be used to hallucinate visually plausible images from noisy or blurry images as shown in Fig. 2.

As discussed, the problem of adding detail to the input image is extremely under-constrained, since there are many high-resolution images that could correspond to the given low-resolution input. This gets worse as the level of magnification is increased, as there is less useful information from the low-resolution image available at large upscaling factors [8]. To address this, we use a large image database to provide us with the "priors" for solving this ill-posed problem. Essentially, we find high-resolution sample images that are similar (but not identical) to the input image and use their information to

---

[1]The term "example-based, single-image hallucination" is the most appropriate name for the approach we describe in this paper, so we shall use the term "image hallucination" throughout.

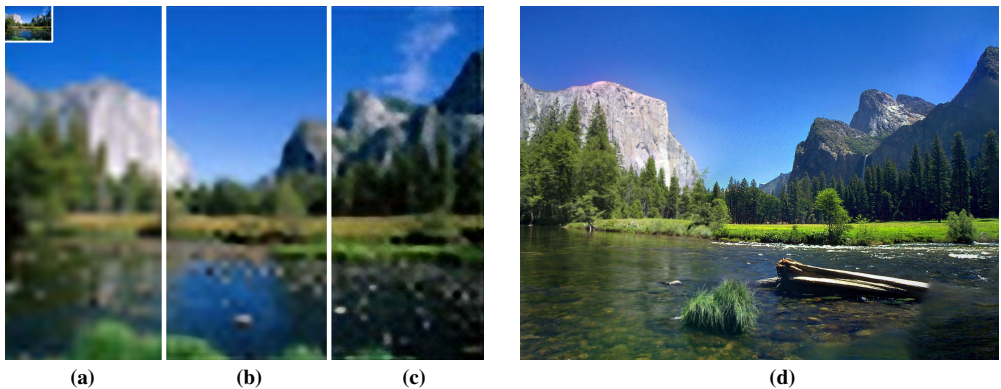

**(a)** **(b)** **(c)** **(d)**

Figure 1: Previous work on image hallucination from a thumbnail image focused on sharpening blurry edges but cannot add new scene detail. For example, the $64 \times 48$ image on the top left is shown upsampled by a factor of $32\times$ with existing techniques: **(a)** bicubic upsampling, **(b)** Timofte et al. [1], **(c)** Schulter et al. [2]. **(d)** Our algorithm, on the other hand, uses an improved patch-based optimization that leverages sample images from a large image database which enables it to synthesize plausible novel detail. Note that the algorithms in (b) and (c) were trained on the same sample images used in (d).

hallucinate plausible detail (see Fig. 3).

Initially, one might wonder whether we could simply use one of the sample images directly instead of trying to inject detail into the upsampled result. This does not work very well, however, as users can see clear differences in composition and pose between the high-resolution sample images and the original input thumbnail, even if it is very small. In fact, our user study in Sec. V-B shows that this naïve approach is actually the least-preferred option among existing methods for the task of image upsampling.

Of course, there has been previous work using image databases for image up-sampling, starting with the seminal work by Freeman et al. [9]. In particular, our work was inspired by the state-of-the-art, "internet scale" super-resolution work of Sun and Hays [6], which uses a patch-based optimization and a very large image database for this purpose. However, their optimization was largely based on the context-constrained optimization of Sun et al. [10] which has problems generating coherent structures, and thus fails to synthesize fundamentally new detail (see Fig. 5).

On the other hand, our optimization is more similar to those of patch-based synthesis approaches (e.g., [11, 12, 13, 14]) which have been extremely successful at solving difficult ill-posed problems, such as filling holes in images. Specifically, these algorithms are able to synthesize large regions of coherent, plausible detail by ensuring the result contains overlapping patches voted from one or more source images. In this paper, we pose our optimization in a similar way to draw high-frequency detail from a large image database.

Furthermore, instead of building an image database from scratch as was done in previous work such as Sun et al. [6], we integrate our algorithm with Google's Image Search feature to tap into Google's extremely large image database, which already contained 10 billion images by 2010 [15]. All together, this new algorithm produces results superior to state-of-the-art approaches at smaller scales, while also synthesizing plausible results at magnification scales never previously demonstrated.

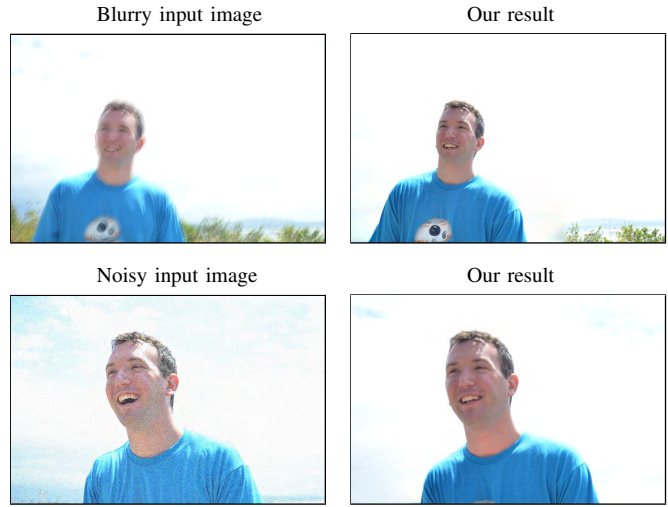

Figure 2: Possible applications of the proposed image hallucination method. Given an input image corrupted by noise or camera-motion blur and similar sample images, the proposed method can hallucinate a visually plausible result by first downsampling the corrupted image to a smaller resolution and then gradually upsampling it.

## II. PREVIOUS WORK

We begin by briefly reviewing the previous work on image super-resolution, image hallucination, patch-based synthesis techniques, and algorithms that use large image databases for image processing tasks.

*A. Image Super-Resolution*

The classical algorithms for image super-resolution (e.g., [16, 17]) take as input a set of low-resolution images of the same scene with small, sub-pixel shifts and then solve for the high-resolution image using the low-resolution measurements as constraints. These *multi-image* super-resolution algorithms are not the focus of this work, since we are interested in situations where there is only one input image. Furthermore, as the magnification factor is increased the low-resolution images provide less useful information for the final result [8].

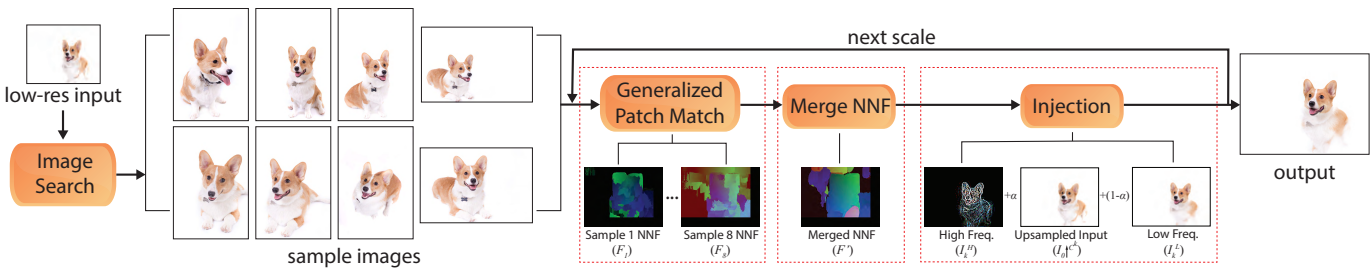

Figure 3: The flow chart of the proposed image hallucination method.

Instead, we are interested in what is known as *single-image super-resolution*. Given that it is a quite important problem, there have been many methods proposed to address it and we only survey some of it here. First, researchers have explored interpolation methods (e.g., [3, 18]), which are straightforward but cannot always sharpen the high-resolution detail. More sophisticated approaches use image statistics or other priors to perform the upsampling (e.g., [4, 19, 20]) and have shown good results in certain cases.

Also related are methods that make a sparsity assumption in the underlying signal in order to perform super-resolution (e.g., [21, 22, 23]). Finally, there are methods that perform special deconvolution to upsample the image (e.g., [24]). However, all of these methods can only sharpen detail that already exists in the low-resolution input and do not synthesize novel detail into the image.

The most powerful approaches available today stem from the seminal *example-based*, single-image super-resolution work of Freeman et al. [5, 9], which implicitly learned features from a database of high- and low-resolution patch pairs of natural images. Effectively, these image patch examples were used as data-driven priors to inject information into the high-frequency pass band of the final image. In their first paper, Freeman et al. [9] used a Markov network to ensure that patches from the database that matched the neighboring regions in the low-resolution input were coherent with each other (i.e., their overlapping pixels were similar). In their second work [5], they extended this approach to a greedy, one-pass algorithm that filled in the high-resolution image in scanline order.

Example-based, single-image super-resolution has been extended in many ways. Some methods were proposed to target specific classes of images (e.g., [25, 26]) which improved results for these cases. Others have added more sophisticated priors or modified the energy equation to give better results (e.g., [27]), changed how the examples are used (e.g., [28]), or improved the low-resolution features so that better matches could be found (e.g., [29]). Others have proposed ways to explicitly learn models that map low resolution images to higher resolution (e.g., [30, 31, 32]), used local linear regressors or support vector regression to upsample the image (e.g., [1, 33, 34, 35]), or used a segmentation of the input into distinct textured regions to improve the reconstruction [10, 36]. More recently, others have proposed using deep learning, such as convolutional neural networks (e.g., [31, 37, 38]), sparse-coding based networks (e.g., [39]), polynomial neural networks (e.g., [40]), recursive networks (e.g., [41]), and generative adversarial networks (e.g., [42]), to learn the mapping between low-resolution and high-resolution images.

### B. Image Hallucination

An example-based, single-image image-hallucination method related to ours is the context-constrained hallucination of Sun et al. [10], which attempts to combine patch-based hallucination with edge smoothness constraints into a single optimization. To do this, the input image is first upsampled naïvely to the target resolution and then segmented into texturally-similar segments. For a given pixel $p$, they use its surrounding texture information to search for 10 similarly textured segments in a universal image database using a filter bank, and then choose the closest patch to $p$'s from these segments. Note that this process is only performed once for each pixel, and this happens *before* the optimization. Furthermore, each pixel searches for patches independently from the other; there is no requirement that the patch selected for one pixel is coherent with the patch selected for its neighbor. During optimization, they solve for the high-resolution image by minimizing an energy equation with a per-pixel hallucination term.

Although this approach has several good ideas, it has some key problems. First, the algorithm only searches for sample patches once, before optimization. Therefore, they cannot leverage new detail that is synthesized during optimization to seek out better patches. Second, there is no guarantee that neighboring pixels find closest example patches that are coherent with each other, which results in the loss of detail. Third, their hallucination energy term forces the optimization to be done at the pixel level (as opposed to over patches), so that each pixel is effectively independent from the others. In this way, one pixel might find detail from one example patch to be the closest, while the neighboring pixel might find it from another. Since no coherency is enforced across the pixels, more detail can be lost. Fourth, the algorithm jumps directly from the low to the high resolution. If this gap is too large, it is very difficult to search for accurate patches in the database. Finally, for extreme magnification the inputs can often be quite small (e.g., $64 \times 48$) and without much texture, making their texture context search less effective.

Later, Sun and Hays [6] extended context-constrained hallucination by improving it in two ways. First, they do not use a universal database of image segments but rather use a larger "internet scale" database to provide them with closer scene matches, and then only use these matches to provide example patches. This improves the quality of the patches they find, plus they have less problems in the transition regions between textured segments. Second, they modify the way candidate patches are selected by taking coherency into account. However, they still perform the same optimization as Sun et al. [10] and so they have many of the same problems.

For this reason, neither of these approaches (nor any of the other example-based image hallucination/super-resolution work for that matter) can synthesize fundamentally new detail into general high-resolution images. This is why was all previous work focused on small magnification factors. Of all the papers we reviewed, the highest magnification shown for a natural, bitmapped image was a factor of $16\times$ [4, 19]. Fattal shows small insets of a child's eye and a portion of an armchair, while Sun et al. shows that they sharpen the stripes on a zebra. In neither case does the algorithm add new detail that does not exist in the original input.

The key difference between many of these previous "patch-based" super-resolution methods (including these last two) and our own is that these previous methods are usually posed as a Markov Random Field since they were inspired by the original work by Freeman et al. [9]. Our approach, on the other hand, is inspired by previous work in patch-based synthesis which uses an alternating minimization to both find good candidate patches while solving for the pixels in the final image. This enables us to add detail in the same way patch-based synthesis has been able to do it when solving other ill-posed problems.

### C. Patch-based Synthesis

Most of the work on patch-based synthesis is based on the seminal work of Wexler et al. [43], Simakov et al. [11], and Barnes et al. [12], which showed how a simple energy formulation enforcing the coherency of patches between a source image and the target could produce plausible results with an alternating optimization. In the first step, the algorithm searches for the closest patches in the source that match the patches in the target. In the second step, these patches are voted (averaged) together to form a new target so that the process can repeat again.

Patch-based optimization like this has been used to successfully fill holes in images [12, 13, 43, 44], morph images [45], retarget images [11], and produce HDR reconstructions [46, 47], in natural and plausible ways.

### D. Image Processing with Large Image Databases

Two key papers that demonstrated the power of large image databases were perhaps the PhotoTourism work of Snavely et al. [48] which used them to reconstruct popular tourist sites, and the work of Hays and Efros [49] which showed how they could tackle difficult ill-posed problems, specifically that of filling large holes in the image with semantically meaningful content. Since then, large image databases have been used to address many image processing problems. Most similar to our approach is the work of Sun and Hays [6] mentioned earlier.

### III. NOVEL PATCH-BASED IMAGE HALLUCINATION

Given an input image $I_0$ of low resolution $w \times h$ and a desired magnification factor $a \gg 1$, the goal of our algorithm is to hallucinate a plausible output image $I_n$ that has much larger resolution $aw \times ah$ and looks like an upsampled version of $I_0$. As noted in the literature, this is an ill-posed problem since there are many high-resolution images that could correspond to the given low-resolution input, particularly when the magnification factor $a$ becomes large.

To tackle this challenging problem, we observe that it can be cast as an optimization that codifies the two key properties of a "successful" output $I_n$. The first property is that when $I_n$ is downsampled, it should be very similar to $I_0$ in an $\mathcal{L}_2$ sense. This ensures that the result looks like it was upsampled from the low-resolution input and is known as the "reconstruction constraint" [8] used in many super-resolution algorithms as part of the energy term (e.g., [10]).

The second property is that the output image should contain new details and other high-frequency information in a semantically plausible way. After all, the low-resolution input contains the low-frequency information of the original image, but its high-frequency information has been lost in the downsampling process [5]. The algorithm should fill in this information in a way that looks natural.

While the first property is fairly easy to satisfy, the second property is more challenging. How can we force the added high-frequency detail to look natural? To answer this question, we turn to the field of patch-based synthesis and use a similar metric: the resulting image will look natural if every patch in the final image is close in an $\mathcal{L}_2$ sense to a real source image. Note that this optimization is subtly, but fundamentally, different than previous patch-based methods that have been used for super-resolution since these first find the closest patches at each pixel and then optimize the final pixel values to match these patches. Since they do not iterate again to refine the closest patches based on the new pixel information, they cannot synthesize new detail that was not present in the original image.

Rather, our algorithm is similar to patch-based synthesis algorithms [13, 43]. In our case, we will continually optimize for both the pixel values and the closest patches in an alternating optimization. To provide good example patches, we will draw the patches from matching images in a large image database. By doing this, our optimization will be able to leverage the database to introduce detail into the upsampled images, thereby producing plausible magnified results.

### A. Framework

We now describe the details of our algorithm. Without loss of generality, the upsampling process can be performed by scaling the image repeatedly by a factor of $c = \sqrt[n]{a}$ and doing this $n$ times. For the optimization to succeed, we note that $c$ must be small (we use $c \approx 1.2$) so that we *gradually* increase the size of the image at every step. This is similar to what is commonly done in other patch-based synthesis approaches, such as image retargetting [11]. We denote the intermediate scaled images as $I_1, I_2, \ldots, I_n$, where $I_k$ is image $I_{k-1}$ scaled by a factor of $c$ and $I_n$ is our final upsampled image. In order to make the final image $I_n$ look plausible and realistic, we draw information from the top $M$ web image matches retrieved via visual search. This gives us sample images $[R_1 \ldots R_M]$, and downsampling these for each scale $k$ gives us candidate images $[C_1^k \ldots C_M^k]$ that are appropriate for synthesis.

The two properties described earlier yield the following energy equation that any good, intermediate upsampled image $I_k$ should minimize:

$$E(I_k) = \alpha(k) \sum_{i \in I_0} ((I_k \downarrow_{c^k})_{(i)} - I_{0_{(i)}})^2 +$$
$$\sum_{p \in I_k} \min_{1 \leq m \leq M} \min_{q \in C_m^k} \left[ d\left(I_k(p), C_m^k(q)\right) + \beta H(p, m) \right], \quad (1)$$

where, in the first term, $I_k \downarrow_{c^k}$ is image $I_k$ downsampled by a factor of $c^k$ to the resolution of $I_0$ (the antialiasing filter is assumed to be included in this operation). This term is the standard reconstruction term [8, 10] that forces $I_k$ to match the low resolution input $I_0$ when downsampled. Note we have weight $\alpha(k)$ on this term to diminish the contribution of this effect as we get to higher resolutions (see Sec. III-G).

The second term ensures that the image $I_k$ looks natural and artifact-free by specifying that every patch around $p$ in $I_k$ should match source patch $q$ in one of the candidate images. Here, $p, q$ are the indices of pixel patches of size $z = 32$ from the intermediate upsampled image $I_k$ and the candidate images, respectively, and functions $I_k(p)$ and $C_m^k(q)$ extract the patches around $p$ and $q$ in these images. Function $d\left(I_k(p), C_m^k(q)\right)$ is the $\mathcal{L}_2$ distance of color and gradient between these patches as computed in the work of Image Melding [13]:

$$d(P, Q) = D(P, Q) + \lambda D(\nabla P, \nabla Q), \quad (2)$$

where each of the $D()$ functions is a standard $\mathcal{L}_2$ distance and $\lambda = 5$. We calculate the distance in the RGB space (unlike Image Melding, which used the L*a*b* color space). $\nabla P$ is the gradient of $P$, and the gradient is calculated by convolving each color channel of P with a kernel $[-1, 1]$ in the $x$ direction and a kernel $[-1, 1]^T$ in the $y$ direction. We search using both color and gradients in order to allow for variations in color tone, lighting, and so on between the current upsampled image and the candidate images.

Finally, the $H(p, m)$ term enforces coherency between the source patches so that we do not get a blurred result in the end by inadvertently averaging together incoherent source patches from different parts of a candidate image or from different candidate images altogether. This expression has two terms:

$$H(p, m) = \alpha_{coh} \text{Coherence}(p, m) + \alpha_{con} \text{Contribution}(m). \quad (3)$$

The first term, given in Eq. 3, ensures that the patches in the neighborhood around $p$ have source patches in the candidate images that are close in position, scale factor, and rotation angle to achieve coherency. The second term, shown in Eq. 3, further improves coherency by limiting the number of candidate images used for source patches. To do this, it computes the score for a source patch from candidate image $m$ using the fraction of the overall upsampled image that has used patches from candidate image $m$. $\alpha_{coh}$ and $\alpha_{con}$ are set to 0.0005 and 0.05 for the results in the paper, and details of *Coherence()* and *Contribution()* are shown in Sec. III-E.

Note that we do not need edge smoothness terms [10] in this equation, because the second term is robust enough to ensure that we are adding coherent, high-frequency detail. Also, unlike some of the earlier patch-based super-resolution work (e.g, [9]), every pixel in image $I_k$ has a patch around it so there is considerable overlap between patches. Finally, we observe that Eq. 1 is similar in form to the energy equations solved in other patch-based synthesis work such as patch-based HDR reconstruction [46], which had one term that enforced the reconstruction to match the sample image when exposed correctly and a second term to transfer information coherently from all the source images. In the next section, we describe how we minimize this equation.

*B. Optimization*

Simultaneously solving for all aspects of Eq. 1 can be difficult because it involves complex inter-relationships between patches, candidate images, the input image, and the output. Therefore, as is common in patch-based synthesis algorithms, we gradually solve for $I_k$ using an alternating minimization where every iteration has four distinct stages, each addressing one aspect of the equation:

**Stage 1 (S1):** In the first stage, for every patch around $p$ in the current version of the target $I_k$, the algorithm finds the best patch around $q$ in each candidate image $C_m^k$ independently. This is done running the generalized PatchMatch acceleration algorithm [50] once for each candidate image, and is equivalent to minimizing the $d(I_k(p), C_m^k(q))$ term in Eq. 1 where the source exctraction $C_m^k(q)$ can handle rotation, scaling, and reflection in addition to translation. Generalized PatchMatch produces a set of $M$ nearest-neighbor fields (NNFs) which indicate the location of the closest patch $q$ in every candidate image for every patch $p$ in $I_k$. Implementation details are in Sec. III-D.

**Stage 2 (S2):** Here the algorithm incorporates the $H(p, m)$ term and the information found in Stage 1 into the patch vote process to minimize the second term of Eq. 1. Before voting, we first compute an *NNF map* which specifies the index of the candidate image with the "best" patch at each pixel location. The best patch is the candidate patch around $q$ with the lowest combined distance $d(I_k(p), C_m^k(q)) + \beta H(p, m)$. This resulting NNF map is smoothed using a majority vote kernel to produce the final NNF map $F'$ which is used in the vote step to vote the candidate patches $C_m^k(q)$ together at each pixel. This stage minimizes the second term of Eq. 1. which ensures that the details added are sharp, coherent, and plausible.

**Stage 3 (S3):** The color and gradient of the source patches are voted using a coherency term to further ensure detail is preserved. The resulting image is reconstructed using the Screened Poisson equation. See Sec. III-E and Sec. III-F for more details.

**Stage 4 (S4):** We now minimize the first term, which ensures that image $I_k$, when downsampled, matches input $I_0$. To do this, we first compute the low and high frequency bands $(I_k^L, I_k^H)$ of the voted $I_k$ with respect to $I_0$'s resolution.

We then alpha-blend the information from $I_0$ into the low-frequency band of $I_k$ to inject this information into the optimization. This ensures that $I_k$, when downsampled, matches the input $I_0$. More details are found in Sec. III-G.

At this point, the new $I_k$ is now the target for the next iteration of the algorithm and this process (stages 1–4) repeats until convergence. Once $I_k$ is finalized, if the target scaling $a$ has not yet been reached then $I_k$ is bilinearly upsampled (to avoid ringing of higher order filters) by a factor of $c$ and the process is repeated again. This relatively simple optimization is all that is needed to upsample a small image to a large scale magnification factor while producing reasonable results. In the subsections that follow, we present more details on various stages of our algorithm, presented as pseudocode in Algorithm 1. A flow chart of our algorithm is shown in Fig. 3.

### C. Computing candidate images

To get appropriate sample images for the search step at each scale, we downsample the original sample images $[R_1 \ldots R_M]$ to get candidate images $[C_1^k \ldots C_M^k]$ which are as close in size to the intermediate image $I_k$ as possible while still maintaining their original aspect ratios. We then compute low-pass versions of the candidate images $[\widetilde{C}_1^k \ldots \widetilde{C}_M^k]$ to use for searching in the first iteration of every scale. We do this because at the first iteration we only have a naïvely upsampled intermediate image $I_k$ which does not yet contain sharp details at its current resolution. By slightly blurring the candidate images for this case, we improve searching. These low-pass candidate images $[\widetilde{C}_1^k \ldots \widetilde{C}_M^k]$ are computed by simply downsampling $[C_1^k \ldots C_M^k]$ by a factor of $1/c$ and then upsampling the result back up by a factor of $c$.

### D. Stage 1: Patch search

In the search step (Sec. III-B, Stage 1), we compute NNFs $[F_1 \ldots F_M]$ for each candidate image independently. As discussed earlier, we use RGB color space and $\mathcal{L}_2$ distance of color and gradient in a generalized PatchMatch [50] framework to get approximate solutions quickly.

To encourage coherency, we use a fixed search radius of 10 from the previous NNF instead of searching over the whole image. This means that, as we improve the NNFs and progress in scale, each patch can only search in a $21 \times 21$ box to look for better matches than the current one. Since there are no NNFs $[F_1 \ldots F_M]$ on which to impose this search constraint at the first iteration of the first scale, we initialize the NNFs at the beginning using SIFT Flow [51]. This method enforces more coherency and image-wide semantics than PatchMatch, preventing erroneous matches between similar-looking things like water and sky.

### E. Stage 2: NNF merge

After the search step has provided NNFs $[F_1 \ldots F_M]$ for each candidate image $[C_1^k \ldots C_M^k]$, we need to merge them into a single NNF that would dictate from which image, and from which position within the image, to get each patch (Sec. III-B, Stage 2). To accomplish this, we compute an *NNF map* $F'$, which stores at every pixel an index specifying which NNF (and thus which candidate image) to use for voting for that pixel's patch (given by the Merge() function in Alg. 1).

---

**Algorithm 1** Patch-based Image Hallucination algorithm

**Input:** Small image $I_0$, magnification factor $a$, search term *keyword* (recommended)

**Output:** Super-resolved, hallucinated image $I_n$

1: /* get sample images */
2: $[R_1 \ldots R_M] \leftarrow \text{GoogleSearchByImage}(I_0, keyword)$
3: /* compute upsample factor c (close to target of 1.2) for exactly $a$ times magnification in $n$ scales */
4: $n \leftarrow \lceil \frac{\log(a)}{\log(1.2)} \rceil$
5: $c \leftarrow \sqrt[n]{a}$
6: **for** scales $k = 1$ to $n$ **do**
7:    $I_k \leftarrow I_{k-1}\uparrow^c$    // bilinear interp.
8:    /* compute candidate images from the samples at same size as $I_k$, low pass filter them for first iteration */
9:    $[C_1^k \ldots C_M^k] \leftarrow \text{ResizeClosest}(\text{size}(I_k), [R_1 \ldots R_M])$
10:   $[\widetilde{C}_1^k \ldots \widetilde{C}_M^k] \leftarrow \text{LowPass}(c, [C_1^k \ldots C_M^k])$
11:   /* upsample NNFs from previous iterations or initialize using SiftFlow in the first scale */
12:   **if** $k == 1$ **then**
13:      $[F_1 \ldots F_M] \leftarrow \text{SiftFlow}(I_k, [C_1^k \ldots C_M^k])$
14:   **else**
15:      $[F_1 \ldots F_M] \leftarrow \text{UpsampleNNFs}([F_1 \ldots F_M], c)$
16:   **end if**
17:   /* injection weight and num. iterations decrease with k */
18:   $\alpha \leftarrow 0.8^{k-1}$
19:   $j \leftarrow \lceil 8 - 7 \cdot \frac{k-1}{n-1} \rceil$
20:   **for** optimization iterations $i = 1$ to $j$ **do**
21:      /* **S1:** compute NNFs by searching for closest patches
22:      **if** $i == 1$ **then**
23:        $[F_1 \ldots F_M] \leftarrow \text{Search}(I_k, [F_1 \ldots F_M], [\widetilde{C}_1^k \ldots \widetilde{C}_M^k])$
24:      **else**
25:        $[F_1 \ldots F_M] \leftarrow \text{Search}(I_k, [F_1 \ldots F_M], [C_1^k \ldots C_M^k])$
26:      **end if**
27:      /* **S2:** compute final NNF map by minimizing second term of Eq. 1 */
28:      $F' \leftarrow \text{Merge}([F_1 \ldots F_M])$
29:      /* **S3:** vote in colors and gradients and reconstruct using Screened Poisson equation */
30:      $[\overline{I_k}, \overline{\nabla I_k}] \leftarrow \text{VoteColorAndGradient}(I_k, F', [F_1 \ldots F_M], [C_1^k \ldots C_M^k])$
31:      $I_k \leftarrow \text{ScreenedPoisson}(\overline{I_k}, \overline{\nabla I_k})$
32:      /* **S4:** blend in input to minimize first term of Eq. 1 */
33:      $I_k \leftarrow \text{Inject}(I_0, I_k, \alpha)$
34:   **end for**
35: **end for**
36: **return** $I_n$

This is computed by first finding, at each pixel, the NNF/candidate image which minimizes the second term of Eq. 1. As discussed, the distance calculation $d(P, Q)$ is given in Eq. 2, and $H(p, m)$ is made up of the two terms in Eq. 3 which are given by:

$$\text{Coherence}(p, m) = -\sum_{i \in \mathcal{N}(p)} \frac{\text{checkCoherence}(F_m(i), F_m(p))}{\text{area}(\mathcal{N}(p))}, \quad (4)$$

$$\text{Contribution}(m) = -\sum_{i \in I_k} \frac{\delta(F'(i'), m)}{w_{I_k} \times h_{I_k}}. \quad (5)$$

In the first term, $\mathcal{N}(p)$ is the neighborhood around $p$ (size $33 \times 33$ in our implementation), while checkCoherence() is a function that checks if two NNF entries are close in every dimension ($x, y$ patch position, scale factor, rotation angle, and reflection). Specifically, this binary function returns a 1 if *all* terms are within a certain threshold of each other (2 pixels are used for position, 0.01 for scale factor, $\pi/20$ for rotation, and the reflection must be identical), and a 0 otherwise. Therefore, this term effectively computes the portion of patches that are coherent in the neighborhood of $p$, and the negative sign decreases the energy when this portion is high.

In the second term, $\delta(a, b)$ is the Kronecker delta function that only equals 1 if $a = b$, and is 0 otherwise, while $w_{I_k}$ and $h_{I_k}$ are the width and height of the target image at current scale $k$ (which has resolution $c^k w \times c^k h$). This term effectively measures the fraction of patches in $I_k$ that come from candidate image $m$, based on the NNF map computed after the previous iteration. Because of the negative sign, this energy term rewards candidate images that provide a large percentage of patches to the image such that, in close cases, the patches will be chosen from the candidate image that has already sourced more patches, which will improve coherency.

As a final step, we smooth the NNF map using a majority vote square kernel that is the same size as the patches. This kernel chooses the sample image $C_m^k$ that appears the most frequently in the kernel window to further enhance coherency.

### F. Stage 3: Vote

At this point, we have the NNFs $[F_1 \ldots F_M]$ for each candidate image $[C_1^k \ldots C_M^k]$ as well as the NNF map $F'$; all three together specify which patch of which candidate image belongs at each pixel in $I_k$. Instead of performing a basic average of the overlapping patches at each pixel, we compute a weighted average with the weight of each patch based on its coherency score which is in the range $[0, 1]$.

This coherency score is computed similar to the one in Eq. 4, as a portion of neighbor patches that are coherent with the current patch. The additional detail here is that two neighboring patches to be voted into $\overline{I_k}$ and $\overline{\nabla I_k}$ are not considered coherent with each other if they come from different sample images. This coherency-weighted vote step provides the last constraint on coherency in order to synthesize sharp details, and pseudocode on this can be found in the supplemental materials. Like the Image Melding method [13], the Screened Poisson equation is then used to solve the reconstructed image using the color and gradient information.

### G. Stage 4: Injection

Once the voted image $I_k$ has been computed, we must alpha blend the input image in order to minimize the first term of Eq. 1. We do this with the following steps:

$$I_k^L \leftarrow (I_k \downarrow_{c^k}) \uparrow^{c^k}$$
$$I_k^H \leftarrow I_k - I_k^L$$
$$I_k \leftarrow I_k^H + \alpha \cdot (I_0 \uparrow^{c^k}) + (1 - \alpha) \cdot I_k^L$$

Effectively, we compute a low-pass version $I_k^L$ of the current image by downsampling it to the size of $I_0$ and then upsampling it back up, and then subtract this from itself to get the high pass version $I_k^H$. We then alpha blend the low frequency version with the original image upsampled to the correct scale and then add the high frequency term back in to get the final result for this iteration.

The blend factor $\alpha$ starts off at 1 at the lowest scale but then decays exponentially at a rate of 0.8 to decrease the injection as the scales progress. This ensures that in early scales, when $I_k$ is close to $I_0$'s resolution, we remain faithful to the details in $I_0$. Later in the scales, when $I_k$ is much larger than the input resolution and the optimization framework has mostly settled on the patch sources, we only partially inject the details of $I_0$ to allow the algorithm to synthesize sharp details.

In the next section, we provide more specific details about our implementation.

## IV. IMPLEMENTATION

### A. Image database and search

To provide the top $M$ sample images $[R_1 \ldots R_M]$ our algorithm needs, we initially experimented with building our own image database and crawled Flickr to create one of a few million images. However, we soon realized that it would be substantially better to leverage Google's "Search by Image" feature and tap into Google's database of over 10 billion indexed images [15]. We did this by writing a Python script with Selenium WebDriver API that integrates this image search directly with our code, and automatically provides us with the $M$ sample images given the input $I_0$.

Note we do not use the "Usage rights" filter as it degrades the search results. The only exception to this is Fig. 4, where we actually show the examples of search results for given queries. In particular, we noticed that using keywords often helps improve the quality of the search results. Although Google will automatically generate a search term for the input image if no keywords are entered, in some cases the generated keyword does not describe the input image very accurately, and so the quality of the super-resolved image decreases due to the lack of good sample images. A full table that includes the keyword used for each test case is shown in the supplemental material. For fairness, we do not include sample images from the internet if they are exactly the same one as the input image.

### B. Color adjustment

Since the sample images from Google Image Search will probably have color differences compared to the input image (due to lighting/shading, white-balancing, or time of day), we perform a color adjustment on the source patches during patch search. We follow HaCohen et al. [52] and Darabi et al. [13] and apply gain $g$ and bias $b$ in the color channels for each source patch to match the target patch. Note that if the color standard deviation of the source patch is too small (less than 5 in [0,255] scale in our experiments), we apply only the bias but not the gain, to avoid a large shift in color. The range of the gain for all the three channels is [1.0, 1.3] and for bias is [-20, 20] in our experiments.

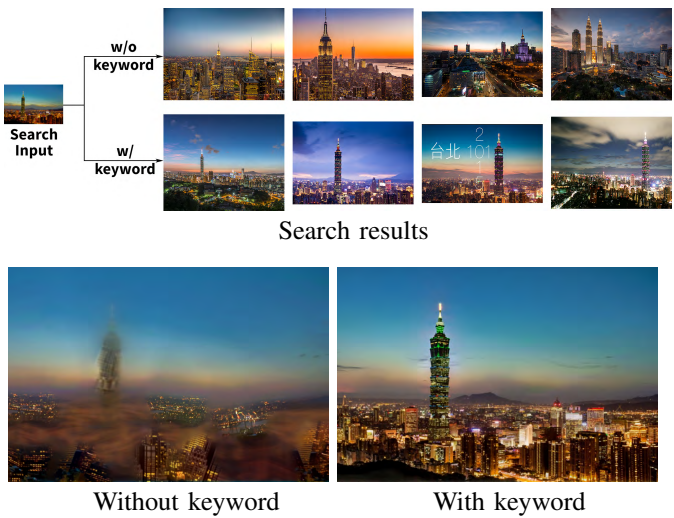

Figure 4: The top row shows the Google Image Search results for the given low-resolution input query. Without a keyword, Google automatically generates the keyword "city," and the search results do not match the input image. However, in many cases the user could provide some information about the image they want to upsample. In this case, the search results with keyword "Taipei 101" match the input scene very well, and our algorithm can use them to generate a visually plausible result. Note that the search results in this (and only this) figure were obtained by setting the "Usage rights" filter to "Labeled for noncommercial reuse" so that we could include the sample images in the paper. These images are queried in May, 2016.

### C. Miscellaneous details

In our algorithm, the number of iterations of the optimization is a linear interpolation of scale, performing 8 iterations at the first scale and 1 iteration at the last scale. This setting was empirically determined to be a good tradeoff between speed and accuracy.

In our experiments, we use $M = 8$ for the number of sample images. This is a good operating point for supplying diversity while avoiding overwhelming the search with too many options. It also has the benefit of efficient parallel computation on quad-core machines with 8 threads, which is what we have used to develop the algorithm. We use a patch size of $32 \times 32$ for all of the experiments. This choice provides a good tradeoff between inserting fine details and maintaining coherency in the final result. Furthermore, it enables memory alignment and allows us to use SIMD acceleration for the algorithm.

## V. RESULTS

The proposed algorithm was implemented in C++ and tested on a variety of input images: natural landscapes, famous landmarks and citiscapes, people, and animals. The images of landmarks were downloaded from Flickr under a Creative Commons license; all other pictures were taken by the authors or their colleagues and are not available online. The only exceptions are, of course, the images of Fig. 5 which are provided by Sun and Hays [6]. After the paper is published, we plan to release a dataset of input thumbnail images and our outputs to allow for comparisons against our approach.

### A. Algorithm Comparisons

To evaluate the performance of our algorithm, in the paper we compare our algorithm against bicubic upsampling as well as several state-of-the-art methods, including some that derive example patches only from the input image [53, 54], patch-based methods that, like ours, draw information from sample images [6, 7, 10], and methods that train on a database of images [2, 34]. For the methods that train on an image database, we trained them using the same sample images used by our technique to optimize their results for each input image.

Furthermore, in the supplemental material we compare our algorithm against 9 additional methods [1, 21, 30, 31, 37, 38, 42, 55, 56] that could not be shown in the paper because of space limitations. Full images of every result, along with the sample images we used are provided there. In all cases we used the code provided by the respective authors, except for two: the algorithm of Yue et al. [7], where the author ran their code for us using our sample images; the algorithm of Ledig et al. [42], where we reimplemented an architecture that works for magnification by a factor of $8\times$. We also run our algorithm at magnification rates that have not been previously demonstrated in the literature for natural, bitmapped images. Examples of a magnification by a factor of $32\times$ are included in the supplemental material.

In Fig. 5, we compare the performance of our algorithm against the "internet-scale" super-resolution approach Sun and Hays [6], as well as Glasner et al.'s "single image" super-resolution [53], and Sun et al.'s "context-constrained" hallucination [10] whose results were provided by Sun and Hays. All images were magnified $8\times$, the maximum published for these previous methods. To ensure that the improvement we see in our results is due to our new algorithm and not the bigger image database, we used 8 out of Sun and Hays's 20 scene matches they provided to reconstruct our final images. As can be seen, in every case our algorithm injected much more detail into the final image and produced more realistic results.

We also tested our algorithm on images of people and animals (Fig. 6), as well as landmarks (Fig. 7). All these results are shown at $8\times$ magnification and compared with the three state-of-the-art techniques [1, 2, 7]. In all cases, our algorithm produces better results than all other methods, a fact confirmed in the user study described below. In terms of timing, our implementation takes about 60 minutes to usample a $128 \times 86$ image to $1024 \times 688$ ($8\times$ magnification).

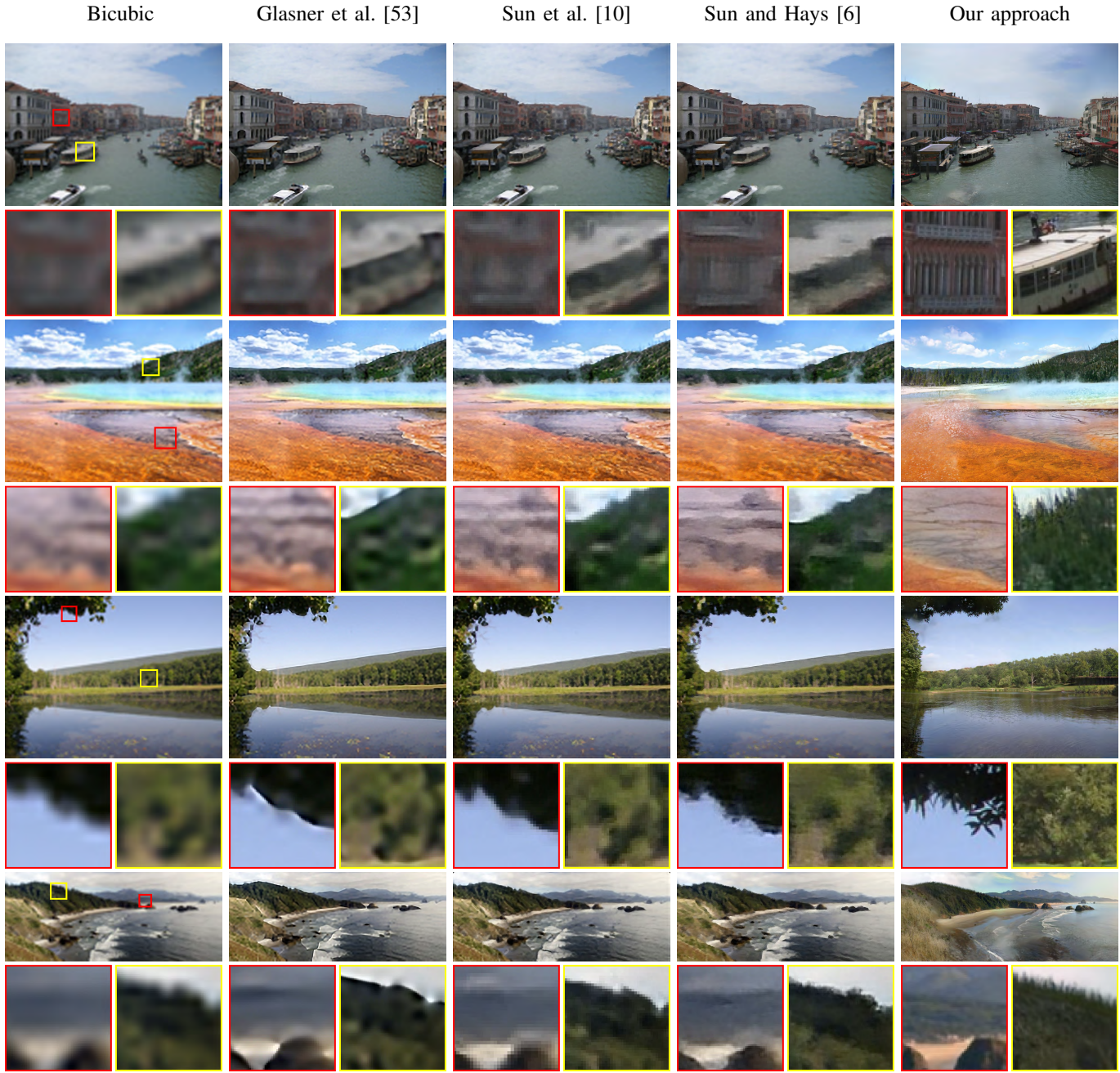

Figure 5: Comparison of our method against three techniques [6, 10, 53], using the input images of Sun and Hays [6]. Our method uses 8 of the matches they provided as sample, not from Google Image Search. All images magnified 8×, the maximum the other algorithms were designed to do.

## B. User Study

To validate our results, we conducted two user studies: the first one asked users to do pairwise comparisons between different results and had 33 participants (26 male and 7 female), aged 21 to 40; the second one asked users to give scores to individual results and had 22 participants (18 male and 4 female), aged 22 to 34. Results of the second user study are shown in the supplemental material. Both studies used a dataset comprised of 24 test images (10 images of people/animals and 14 images of landmarks) and had a total of four tasks, the order of which was randomized to reduce training bias.

To conduct the pairwise user study, we created a website that allows each subject to toggle between two candidate images and see the differences. For a set of images, four tasks were evaluated: **(1)** their realism (could the images fool someone into thinking they were real images captured with an actual camera), **(2)** their visual quality (they did not have objectionable artifacts), and **(3)** their closeness to the input thumbnail (the image had to resemble the thumbnail when downsampled), **(4)** both their visual quality and closeness to the input thumbnail at the same time. Each user was asked to choose from the pair of candidate images the one that most closely fulfilled the description of the task. This comparison was done for upsampled results produced by bicubic interpolation, our algorithm, and four state-of-the-art super-resolution algorithms [1, 2, 7, 54], as well as the sample images themselves (for a total of 7 different "methods"). The sample image used in study was the one that sourced the most patches during synthesis (i.e., provided the most information for our final result). Although the total number of pairwise comparisons between the different methods is $\binom{7}{2} = 21$ for each scene, for the sake of time we asked the users to compare 50 pairs drawn from 10 scenes (5 pairs/scene) in each task of the user study.

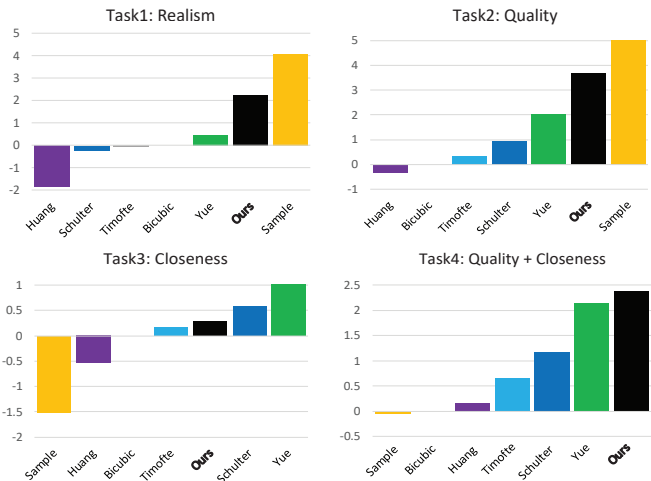

Figure 8: Our user study performs pairwise comparisons between our algorithm and the sample image, a bicubic-upsampled image, and four state-of-the-art super-resolution methods [1, 2, 7, 54].

After gathering the data, we then used Bradley-Terry model [57] to do the global ranking for pairwise comparisons. Given the count of a pairwise comparison $i > j$, where $i$ and $j$ are two different methods, the model estimates the probability of method $i$ is better than method $j$: $P(i > j) = \frac{e^{s_i}}{e^{s_i}+e^{s_j}}$, where $e^{s_i}$ and $e^{s_j}$ are calculated by maximum likelihood estimate using the pair comparison results, $s_i$ and $s_j$ are the B-T scores of method $i$ and $j$.

The results of our user study are shown in Fig. 8. The charts show the B-T scores of each method, and scores of all methods are standardized so that Bicubic has a score of 0. We see that our algorithm outperforms all four competing algorithms, in terms of realism and quality. When comparing our method to next-best other method on these tests [7], the probability that our method is prefered by the users is 85.59% for realism, and 84.06% for quality. In the third task, our method is slightly worse than some other methods in terms of closeness to the thumbnail, but the sample image is much worse than all the others. This means that users can easily tell the difference between the sample image and the thumbnail. In the third task of the second user study shown in the supplemental material, it shows that when users see only one image at a time, our method is comparable to the others in terms of closeness to the thumbnail. Finally, in the last task our method was preferred over the others, and the sample image is the least preferable among all candidates. From the results of the third task and fourth task, we know that the sample image would not be a good substitute for upsampling the image.

*C. Limitations and future work*

Although our results demonstrate a considerable improvement over existing approaches, there are still many limitations to the proposed approach. First of all, our algorithm works best for larger magnification factors. When the upsampling rate is very small (e.g., less than $4\times$) it may make more sense to use other methods, such as the statistical priors (e.g., [4]), since they work better and produce more predictable artifacts.

Furthermore, our method relies on finding good image matches to produce reasonable results. Even with Google's large image database there are sometimes problems with the candidate images, which causes the injected detail to be semantically incorrect. Some of these artifacts can be seen in the images in this paper and in the supplementary material. Of course, this problem reduces with bigger image databases, and companies like Google have been growing their database steadily over the past decade. This could also be addressed by doing localized searches on subsets of the image, like has been done in other work (e.g., [10]). This would allow us to find better, localized detail from specific sources, as opposed to searching for entire images that match the input. There could also be ways to allow the user to provide more semantic context for the images. For example, it would be interesting to combine our algorithm with sketch-based synthesis algorithms like Sketch2Photo [58] to produce more realistic textures.

## VI. CONCLUSION

In this paper, we presented a novel, patch-based algorithm for example-based, single-image image hallucination, which allows an input image to be upsampled up to $32\times$ its original size. Our algorithm is based on an optimization which codifies the two key objectives of image hallucination for super-resolution: to produce an upsampled image that (1) matches the low-resolution image when downsampled, and (2) contains natural, high-frequency detail. To address the second objective, we use an energy term similar to previous work in patch-based synthesis, which specifies that the reconstructed image will look natural if every patch in it can be found in source images. We can leverage Google's image database to provide these candidate images or use a personal image collection. This work is only the first demonstration of the feasibility of upscaling with large mangification factors, and expect that future work in this area will lead to further improvement.

## ACKNOWLEDGMENTS

The authors would like to thank Chris Sweeney for initial discussions, Abhishek Badki for discussions and helping to run some of the competing algorithms, and all the volunteers for participating in the user study. This work was partially funded by National Science Foundation grants IIS-1342931, IIS-1321168, and IIS-1619376.

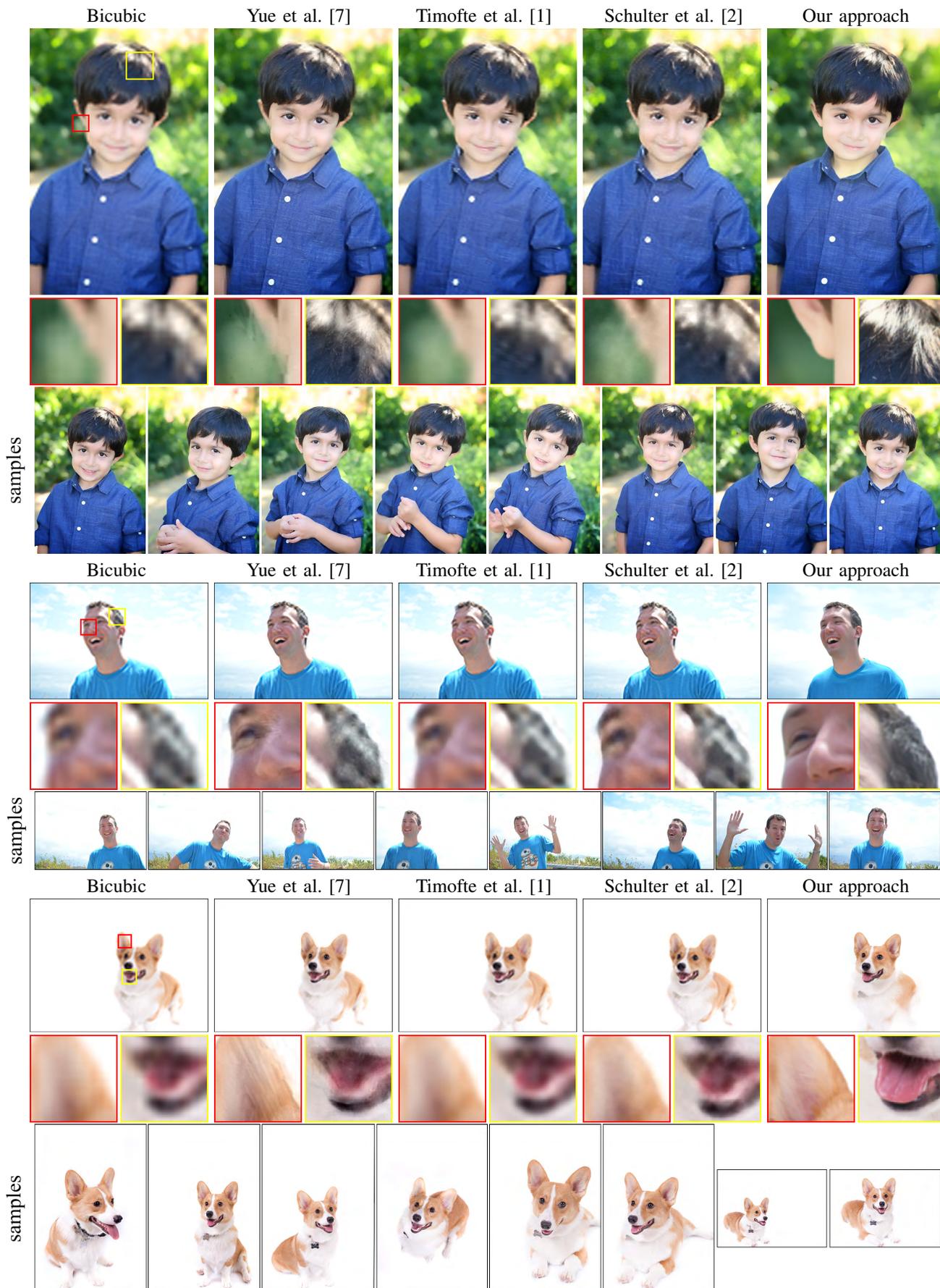

Figure 6: Comparison of our algorithm against three state-of-the-art techniques [1, 2, 7] on images of people or animals. Both the samples and input image are from personal image collections. The third row of each scene comparison are the sample images we used for all algorithms (except for bicubic, of course). All images are magnified by $8\times$ in each dimension, with the maximum dimension going from 128 to 1024 pixels. The full images (along with the sample images used) can be found in the supplemental material.

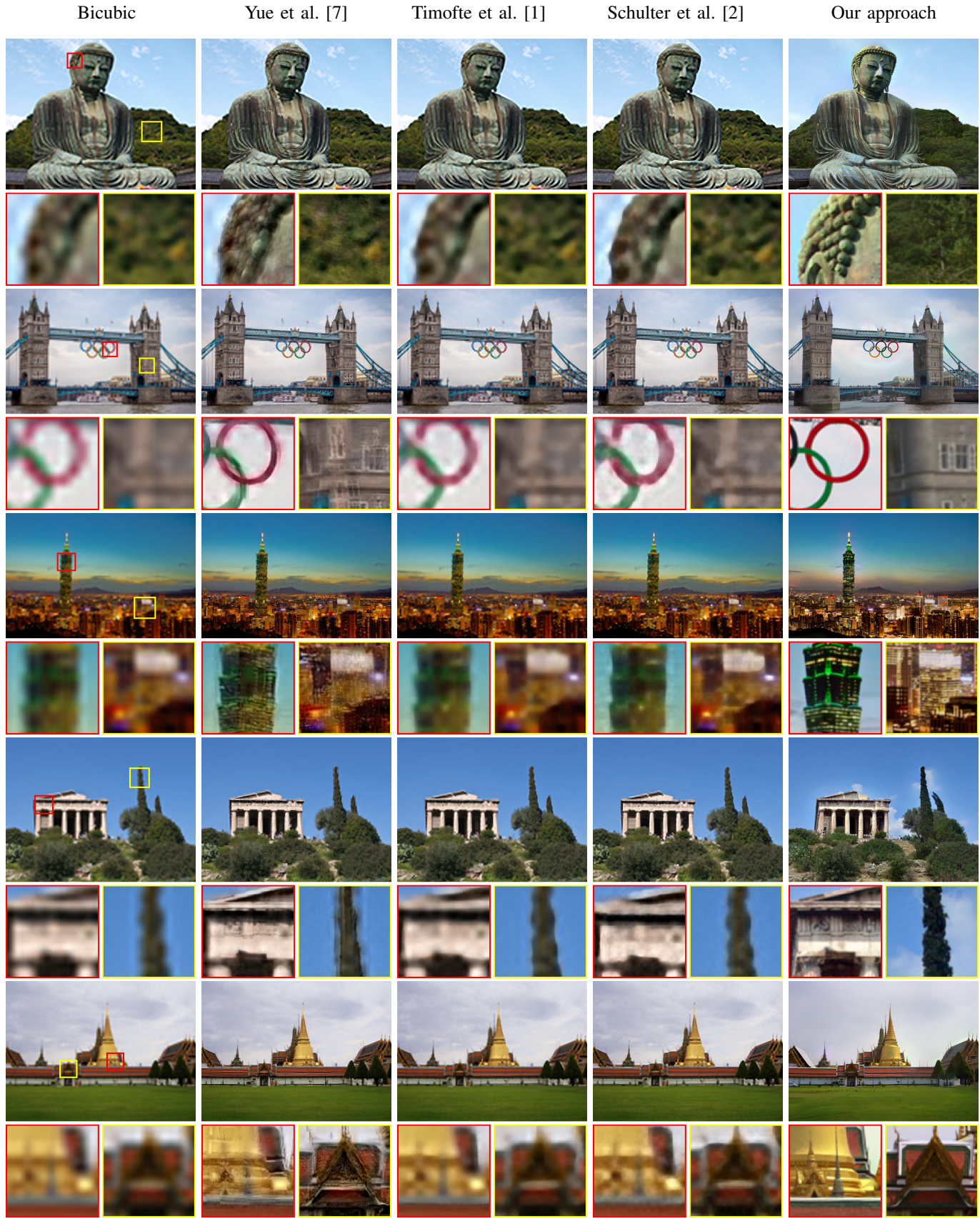

Figure 7: Comparison of our algorithm against three state-of-the-art techniques [1, 2, 7] on images of landmarks downloaded from Flickr, with sample images from Google Image Search. All images magnified by $8\times$ in each dimension, with the maximum dimension going from $128$ to $1024$ pixels. Note that all methods (except for bicubic, of course) leverage the same sample images our method used. The full images (along with the sample images used) can be found in the supplemental material.